\documentclass{article}
\usepackage{spconf,amsmath,graphicx}

\title{SPEECH-LANGUAGE PRE-TRAINING FOR END-TO-END SPOKEN LANGUAGE UNDERSTANDING}
\name{Yao Qian$^1$, Ximo Bian$^2$\sthanks{Work done while the second author was an intern at Microsoft.}, Yu Shi$^1$, Naoyuki Kanda$^1$, Leo Shen$^1$, Zhen Xiao$^1$  and  Michael Zeng$^1$}

\address{
  $^1$Microsoft Cognitive Services Research Group, USA,  
  $^2$Beijing Institute of Technology, China }

\begin{document}
\ninept
\maketitle
\begin{abstract}
End-to-end (E2E) spoken language understanding (SLU) can infer semantics directly from speech signal without cascading an automatic speech recognizer (ASR) with a natural language understanding (NLU) module. However, paired utterance recordings and corresponding semantics may not always be available or sufficient to train an E2E SLU model in a real production environment. In this paper, we propose to unify a well-optimized E2E ASR encoder (speech) and a pre-trained language model encoder (language) into a transformer decoder. The unified speech-language pre-trained model (SLP) is continually enhanced on limited labeled data from a target domain by using a conditional masked language model (MLM) objective, and thus can effectively generate a sequence of intent, slot type, and slot value for given input speech in the inference. The experimental results on two public corpora show that our approach to E2E SLU is superior to the conventional cascaded method. It also outperforms the present state-of-the-art approaches to E2E SLU with much less paired data.
\end{abstract}
\section{Introduction}
\label{sec:intro}

The conventional SLU generally contains two components: the automatic speech recognizer (ASR) to decode the input speech into recognized text and the natural language understanding (NLU) module to transform the ASR hypothesis into a concept or semantic label that can drive subsequent responses in a task-oriented spoken dialog system (SDS). These two components are optimized with different criteria. Nowadays, the performance of ASR has been significantly improved by using a huge amount of training data with a deep learning architecture. Transfer learning based on transformer-based LMs like BERT \cite{DBLP:journals/corr/abs-1810-04805} or GPT \cite{radford2019language} has also largely boosted the accuracy of NLU. This cascaded process is beneficial in reusing the well-optimized modules to accelerate the development of the SLU system. However, the cascaded process could propagate errors that occurred in the current module to the following modules and results in generating inappropriate responses or even causing the task failure.   

In recent few years, an end-to-end (E2E) modeling approach has been widely applied to speech recognition, language recognition, etc. It enables us to utilize as little a prior knowledge as possible and train a single model for the whole target system, skipping the intermediate modules in conventional pipeline designs. ASR-free E2E SLU has also been exploited in \cite{qian2017exploring,chen2018spoken,haghani2018audio,lugosch2019speech,serdyuk2018towards,9053163}, where either the raw waveforms or the acoustic features (e.g., filterbank features) are directly used as the inputs of SLU to infer semantic meanings. In \cite{haghani2018audio}, multiple E2E SLU approaches, i.e., direct model, joint model, multi-task model, and multi-stage model, have been proposed to jointly optimize the ASR and NLU components in a manner of sequence-to-sequence mapping. On the other hand, the performance of the end-to-end approach is always suffering from a dearth of data, i.e., it is difficult to obtain large amounts of paired speech utterances and corresponding semantic labels from real production environments when we are developing new SDS applications from scratch. 



Substantial work has shown that transfer learning with a pre-trained model requires much less task-specific data than conventional methods. It is also extended to cross-modal learning, e.g., VLP \cite{zhou2019unified} and SpeechBERT \cite{chuang2020speechbert} proposed to build two-modality joint embeddings for image captioning, spoken question answering, etc. There are many studies to use either speech model or language model as a pre-trained model to address the issue of insufficient data for SLU but few tries to leverage both and jointly optimize them on the paired data. BERT or GPT is leveraged as the pre-trained language model \cite{huang-chen-2020-learning,yqian_interspeech2020}. The speech model is pre-trained by using ASR task \cite{lugosch2019speech,9053163,9053281}, i.e., to predict phonemes, sub-words or words, or by self-supervised learning on the unlabeled speech data \cite{chung2020semisupervised}. Language embeddings were assumed to have better representation for lexical semantics than speech embeddings, so a loss function, e.g., mean squared error (MSE), L1 norm, or triplet loss, is employed to make speech embeddings closer to language embeddings in the stage of jointly training semantic representations across these two modalities \cite{9053281,chung2020semisupervised}. This kind of alignment for speech and language embeddings could be applied at the sentence level or token level. Sentence-level alignment is more feasible than that of token-level since obtaining accurate boundaries of sub-word for speech embedding sequence is nontrivial. Consequently, most of the studies focus on sentence-level SLU, i.e., domain/intent classification, or using a fixed dimension vector of logits corresponding to the different semantic labels.

In this paper, we exploit leveraging transfer learning based on well-trained encoders from both speech and language for sentence-level and word-level SLU, i.e., intent classification and slot filling. The proposed approach can compensate for insufficient paired training data in the SLU tasks and avoid train a model from scratch so that the resultant model has a better generalization capability. 
The major contribution of this paper is two-fold:
\begin{enumerate}
\item	We unify two encoders from speech and language separately into a decoder, continually pre-train it by using a conditional masked language model (MLM) objective, and finetune it to generate a sequence of intent, slot type, and slot value.
\item	Slot filling, conventionally regarded as a classification or tagging problem on the word/token level, is formulated as a conditional sequence to sequence generation problem, given the contextual embeddings of input speech. The effectiveness of this approach is demonstrated by a comparison with the conventional method on public datasets.  

\end{enumerate}

\section{Related Work}
\label{sec:RelatedWork}

Inspired by the transfer Learning with a unified text-to-text transformer (T5) \cite{raffel2019exploring} where the input and output are always text strings for NLP tasks,  fine-tuning a pre-trained causal  LM (e.g., GPT), for NLP tasks like natural language understanding, natural language generation, or end-to-end task completion have been explored for dialog systems \cite{DBLP:journals/corr/abs-1902-10909,peng2020few,wu2019transferable,budzianowski2019hello}. However, most of them aim at processing the user’s written responses or the human transcriptions of spoken responses. Speech to semantics mapping was also defined as a sequence-to-sequence problem in \cite{haghani2018audio,9053063}, but it lacks a mechanism to leverage pre-trained models from either speech or language to further improve the performance of E2E SLU. Speech synthesis is thus explored to generate large training data from multiple artificial speakers to cover the shortage of paired data in E2E SLU \cite{9053281,9053063}.

The most similar work to ours has been presented in \cite{9053281,denisov2020pretrained}, where it initializes the speech-to-intent (S2I) model with an ASR model and improves it by leveraging BERT and text-to-speech augmented S2I data. The significant difference between ours and theirs is that they employ sentence-level embedding from both speech and language to jointly classify intent, while we use an attention-based autoregressive generation model jointly trained by using frame-level speech embedding and token level text embedding to generate multiple intents and slots/values given an utterance input. It is difficult to use their approach to handle more complicated SLU tasks, e.g., slot filling in the ATIS dataset, since a classification layer with a fixed-length output is impossible to cover various slots/values across utterances. 

\section{SPEECH-LANGUAGE PRE-TRAINING FOR SLU}
\label{sec:PST_SLU}

\begin{figure}[htb]

\begin{minipage}[b]{1.0\linewidth}
  \centering
  \centerline{\includegraphics[width=8cm]{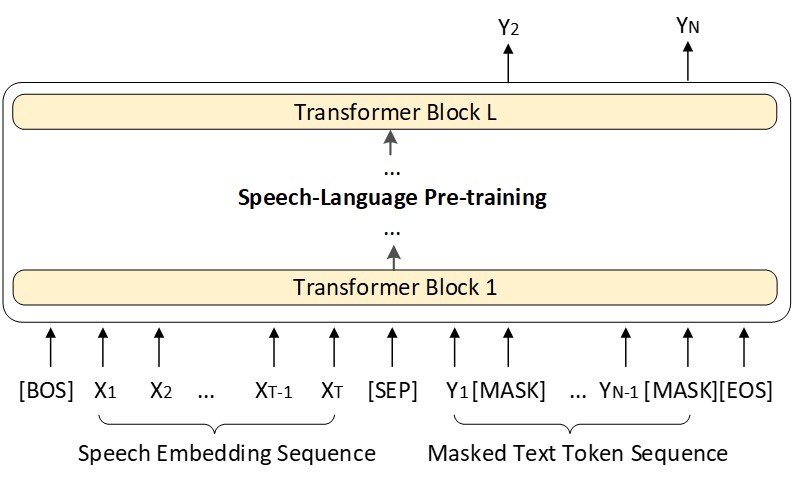}}
 \vspace{-0.3cm}
\end{minipage}
\caption{A schematic diagram of our approach to speech-language pre-training for E2E SLU.}
\label{fig:diagram}
\end{figure}

Our proposed model architecture for speech-language pre-training is depicted in Figure~\ref{fig:diagram}. The model inputs include speech embedding sequence, $X$, masked text token sequence, $Y$, and three special tokens. Speech embeddings are extracted from the encoder of an attention-based encoder-decoder ASR model. We tokenize the texts into subword units by WordPiece \cite{DBLP:journals/corr/WuSCLNMKCGMKSJL16}, and add the corresponding token embeddings with position embeddings. Three special tokens: [BOS], [SEP], and [EOS] indicate the start of speech input, the boundary between the speech input and the text input, and the end of text input, separately. These three delimiter tokens only give the model weak information about which segment (i.e., speech input or text input) each embedding belongs to. We add the segment embeddings to both speech embeddings and text embeddings. 

\subsection{Pre-training}
\label{ssec:pre-training}
We initialize the parameters of the model with pre-trained BERT and continually pre-train it with the speech and the corresponding transcriptions from the target dataset. The parameters of the model are updated by using a cloze task with a conditional masked language modeling (C-MLM) objective. It aims to predict the masked token $y^m_n$ in $Y = (y_1,..,y_N)$, conditioned on a sequence of input $X$. It is usually trained by Maximum Likelihood Estimation (MLE), or equivalently, minimizing the cross-entropy loss as follows:
\begin{equation}
\mathcal{L(\theta)} = -\sum_m \log P_\theta(y^m_t|y_{1:t-1},X) 
\label{eq1}
\vspace{-2pt}
\end{equation}
where the conditional probability is calculated by using an attention mechanism in transformer-based network architecture. 

We randomly select some tokens in the input and replace each of them with a special token [Mask] or a random token or the original token, and then predict the masked tokens by feeding the output vectors from the transformer network into a softmax classifier. We mask one token or a bigram/trigram at random. For the prediction, the tokens in the speech part can attend to each other, while the tokens in the text part can only attend to the left-side context (including all the tokens in the speech part) and itself. It is different from what BERT uses, i.e., all tokens are allowed to be attended in the prediction.

A self-attention mask matrix, as defined in \cite{NIPS2019_9464},  
\begin{equation}
M_{ij}=\left\{ 
        \begin{array}{lll}
         0, && \text{allow to attend} \\ 
         -\infty, && \text{prevent from attending}  
         \end{array}
         \right.
\label{eq2}
\vspace{-2pt}
\end{equation}
is applied to control which contextual tokens can be attended to. The $l$-th self-attention output, $A_l$, is computed as:
\begin{equation}
A_l = \text{softmax}(\frac{QK^\top}{\sqrt{d_k}}+M)V
\label{eq3}
\vspace{-2pt}
\end{equation}
where we use a single attention head in the self-attention module to simplify the description. $V$, $Q$
and $K$ are the intermediate variables as defined in \cite{NIPS2017_7181} for representing values, queries, and keys, respectively. ${d_k}$ is the dimension of the keys. 

Our continual pre-training approach can be regarded as turning the BERT initialized encoder, which has learned better text representations by encoding contextual information from both directions, into an autoregressive decoder that satisfies the property of sequence-to-sequence generation in the inference stage. 

\subsection{Fine-tuning for SLU}

We fine-tune the pre-trained speech-language model by using the same objective as that of pre-training but with the speech and the annotated semantic labels from the target dataset. Unlike the conventional slot IOB labels (i.e., tagging inside, outside, beginning words in a slot type ) assigned to every word in the utterance, sequence representation, such as [intent] \& [slot1 type] [slot1 value] \& [slot2 type] [slot2 value] \& ..., is employed to represent the intents and the slots/values for the whole utterance.  

\subsection{Inference}

In the inference stage, we first encode the speech embeddings along with the special [BOS] and [SEP] tokens and then use a greedy sampling approach or a beam search to generate text token-by-token. The generation starts with appending a [MASK] token to the input sequence, replaces it with a sampled token, and repeats the process till the [EOS] token is chosen. Given speech embeddings, the pre-trained speech-language model can directly generate the corresponding transcription and the fine-tuned model can produce the matching semantic label sequence. The model can also generate ASR transcription and SLU semantic labels sequentially if the input text token in Figure \ref{fig:diagram} is the concatenation of transcription and semantic labels during the training. 

\section{EXPERIMENTS AND RESULTS}
\label{sec:experiments}

\subsection{Datasets}
\label{ssec:Datasets}
We evaluate our approach of speech-language pre-training to E2E SLU on two public SLU datasets: Fluent Speech Commands (FSC) \cite{lugosch2019speech} and Air Travel Information System (ATIS) \cite{10008827816}.

FSC dataset was recorded by 77 speakers for a smart home or virtual assistant. There are 248 distinct sentences.  Each sentence, e.g.,\{turn on the lights in the kitchen\}, was spoken by multiple speakers in both training set and validation/test sets. Each audio file is labeled with three slots, e.g., \{action: “active”, object: “lights”, location: “kitchen”\}. We follow the usage suggested in \cite{lugosch2019speech}, i.e., combining slot values as the intent of utterance and using 31 distinct intents in total.

ATIS pilot corpus is the most commonly used dataset for SLU research. It comprises the acoustic speech data for a query, the transcriptions of that query, and the corresponding semantic frames, i.e., an intent and slots filled with phrases. For example, Query: \{What flights are available from Pittsburgh to Baltimore on Thursday morning\}; Intent: \{flight info\}; Slots: \{\texttt{from\_city}: “Pittsburgh”, \texttt{to\_city}: “Baltimore”, \texttt{depart\_date}: “Thursday”, \texttt{depart\_time}: “morning”\}

The statistics of these two datasets are shown in Table~\ref{tab:data}, where lists the number of audio files in training, validation and test sets, and the number of intents and slots. We use the same data division as \cite{lugosch2019speech} and \cite{goo-etal-2018-slot,DBLP:journals/corr/abs-1902-10909} for FSC and ATIS datasets, respectively. Some utterances have multiple intents in ATIS corpus, so we employed 21 intents rather than 18 intents in some papers, where only one intent was forcedly assigned to an utterance. Slot labels in the IOB format (120 labels in total) and human transcriptions have been widely used in the studies (e.g., \cite{DBLP:journals/corr/abs-1902-10909}) for ATIS corpus. We find 39 utterances have the transcriptions, but the audio files are missing. We use total of 4,439 training utterances and the corresponding 79 original slot types in this study.

\begin{table}
\caption{Number of audio files in training, validation and test sets, and the number of intents and slots}
\smallskip
\renewcommand{\arraystretch}{1.3}
\label{tab:data}       
\centering
\begin{tabular}{cccccc}
\hline \hline\noalign{\smallskip}
\# & Training & Validation & Testing & Intents & Slots \\ 
\noalign{\smallskip}\hline
FSC & 23,132 & 3,118 & 3,793 & 31 &  \\ 
ATIS & 4,439 & 500 & 893 & 21 & 79 \\ 
\noalign{\smallskip}\hline\hline
\end{tabular}
\end{table}

\begin{table*}
\centering
\caption{Performance(\%) of different systems}
\smallskip
\renewcommand{\arraystretch}{1.3}
\label{tab:performance}       
\begin{tabular}{cc|cc|ccccc}
\hline \hline\noalign{\smallskip}
\multicolumn{2}{c|}{Systems} & \multicolumn{2}{c|}{FSC } & \multicolumn{5}{c}{ATIS}\\
 Approaches & Configurations  & WER & Intent-Acc & WER & Intent-Acc & Slot-Pre & Slot-Rec & Slot-F1  \\
\noalign{\smallskip}\hline
E2E Joint & $ASR\cup SLU$   & 0.86    & 99.13     &  & &&   & \\
\noalign{\smallskip}\hline\noalign{\smallskip}
E2E SLP & one-step ($SLU$) &    &    98.58 &     & 94.96 & 89.21 & 85.58 & 87.36\\
E2E SLP & one-step ($ASR\cup SLU$) & 0.42   & \textbf{99.71}   & 14.47 & 95.30 & 86.71 & 82.20 & 84.40 \\
E2E SLP & two-step ($ASR, SLU$) & 0.42   &  \textbf{99.71}  &   8.68    &  \textbf{96.30} & \textbf{91.13}  & \textbf{90.76} & \textbf{90.95}\\
\noalign{\smallskip}\hline
Cascaded & Ref $\rightarrow$ BERT  &    &  100.0   &  & 97.65 & 95.89 & 96.30 & 96.10 \\
Cascaded & ASR $\rightarrow$ BERT   & 2.82  & 96.84   & 14.82 & 95.18 & 86.53  & 81.86 & 84.13 \\
Cascaded & SLP ASR $\rightarrow$ BERT  & 0.42 &  99.47  &  8.68 & 95.52 & 90.46 & 89.78 & 90.12 \\
\noalign{\smallskip}\hline\hline
\end{tabular}
\end{table*}

\subsection{Experimental setup}
\label{ssec:Setup}

The experiments are configured as follows:

\textit{\textbf{E2E with speech-language pre-training (E2E SLP)}} This is our approach described in Section~\ref{sec:PST_SLU}. We construct mode with the architecture shown in Figure~\ref{fig:diagram}. The model parameters are initialized by using the English uncased BERT-Base model, which is a multi-layer bidirectional transformer encoder consisting of 12 layers, 768 hidden states, and 12 heads, and is trained on  BooksCorpus (800M words) and English Wikipedia (2,500M words). 

The input to the model is a concatenation of text embeddings, speech embeddings, and three special tokens: [BOS], [SEP], and [EOS]. Text embeddings are the sums of WordPiece embeddings with a 30,000 token vocabulary, positional embeddings, and segment embeddings. Speech embeddings are extracted from the outputs of the encoder of an attention-based encoder-decoder (AED) ASR model, which comprises of 6 layers of 1,024-dim bidirectional long short-term memory (BLSTM) for the encoder, 2 layers of 1,024-dim unidirectional LSTM for the decoder, and a conventional location-aware content-based attention \cite{NIPS2015_5847} with a single head in between the encoder and the decoder. Layer normalization \cite{ba2016layer} was applied after every BLSTM in the encoder but not after every LSTM in the decoder. The input feature is a sequence of 80-dim log Mel filter bank with a stride of 10 milliseconds (ms). Three of them are stacked together to form a 240-dim super-frame. We feed into the encoder on top of the stacked features. 4,000 WordPiece tokens are used as the output targets. This model was trained on 75 thousand hours of Microsoft anonymized training data. A linear layer is applied to the speech embeddings for converting 1,024-dim to 768-dim in order to keep the same dimension as those of text embeddings. 

 Adam optimizer, 64 batch size, and 1e-4 learning rate are used for both pre-training and fine-tuning. We trim long utterance and pad short utterances to 30 words for the text embeddings and 500 frames for the speech embeddings. During the training, 15\% of input text tokens are masked and replaced with the special token [Mask], the random token, or the original token with the probabilities of 80\%, 10\%, and 10\%, separately. One or two/three successive tokens are masked at random with the chances of 80\% and 20\%, respectively. A beam search with beam size 4 is used in the inference stage.

\textit{\textbf{E2E Joint Modeling}}   Similar to the AED-based ASR model, the approach of E2E joint modeling to SLU also employs an encoder-attention-decoder architecture and trains the model in an end-to-end manner. We use the same architecture and input acoustic features as those of the above AED-based ASR model to train E2E SLU. The output of this model is the concatenation of transcription and its corresponding semantics rather than just the transcription, i.e., the decoder produces the transcription followed by intent, slot types, and slot values. We test this approach by using only the FSC corpus. The training data in ATIS is too insufficient to deliver a decent E2E model. 

\textit{\textbf{Cascaded ASR+NLU}}  It is a conventional approach to SLU. We first use ASR described in the above section to recognize the spoken input into text and then fine-tune the pre-trained BERT model with the ASR transcriptions of training utterances. The objective of the fine-tuning is to jointly optimize model parameters for intent classification and slot filling \cite{DBLP:journals/corr/abs-1902-10909}. This approach can leverage BERT to improve the generalization capability of NLU models and achieve the state-of-the-art performance on the human transcriptions of ATIS corpus. The same English uncased BERT-Base model as that employed in our approach is used for fine-tuning. The hyperparameters are also set as those in \cite{DBLP:journals/corr/abs-1902-10909}.

\subsection{Results and discussion}
\label{ssec:restuls}
Word error rate (WER) is used to evaluate the performance of ASR. The performance metrics for semantic labeling are intent accuracy (Intent-Acc), slot precision (Slot-Pre), slot recall (Slot-Rec), and slot F1 (Slot-F1). Table~\ref{tab:performance} shows the performance of different systems. The slots/values of each utterance are combined as the intent of the utterance in the FSC corpus, so there are no results related to the slots in Table~\ref{tab:performance}. Meanwhile, the results of E2E joint modeling for ATIS corpus are also blank due to the inadequate training data for building an E2E SLU model. 

The E2E joint modeling approach to SLU achieves the intent accuracy of 99.13\% on the FSC testing set. This is the best performance we have seen by building a model from scratch, i.e., using only training data of FSC without any helps from the pre-trained models.

We apply two strategies in building E2E SLP modes. Two-step means that we firstly pre-train SLP with paired speech and the corresponding transcriptions, and then fine-tune the resultant model with paired speech and the semantic labels. One-step indicates that we just fine-tune the SLP model. The output targets can be semantic labels, i.e., one-step ($SLU$), or the concatenation of transcriptions and semantic labels, i.e., one-step ($ASR \cup SLU$). The experimental results in Table~\ref{tab:performance} show that the performance of one-step ($ASR \cup SLU$) is equivalent to that of two-step ($ASR, SLU$) for the FSC corpus but two-step significantly outperforms one-step for the ATIS corpus. We think 1) the size of training data in FSC is much larger than that in ATIS so that both strategies can converge to the same performance; 2) the concatenations of transcriptions and semantic labels in ATIS corpus are much longer than those in FSC corpus. It may cause performance degradation for sequence generation. 

Three kinds of transcription: 1) The ground-truth transcription (Ref$\rightarrow$BERT) as a upper-bound reference; 2) ASR transcription (ASR$\rightarrow$BERT) by using the AED-based ASR system mentioned in Section~\ref{ssec:Setup}; 3) ASR transcription from pre-trained SLP model (SLP ASR$\rightarrow$BERT), are fed into the BERT model to build cascaded SLU system. The pre-trained SLP model can reduce ASR WER from 2.82\%, 14.82\% to 0.42\%, 8.68\% for FSC and ATIS, respectively.  The refined ASR hypotheses are critical for improving the performance of intent prediction and slot filing, e.g., the slot-F1 for ATIS is significantly improved from 84.13\% to 90.12\%. In the pre-training stage, we continually train the SLP model with the speech embeddings from the encoder of the AED-based ASR model and the tokenized transcription by using FSC or ATIS dataset. This procedure can be thought of as domain adaptation for the acoustic model (AM) and language model (LM) in the point of view of ASR customization. LM represented by the decoder in the AED-based ASR model is replaced by BERT based LM and further adapted with the training transcriptions of FSC or ATIS. It brings most of the WER reductions for the generated hypotheses. 

By comparing the performance of different systems shown in Table~\ref{tab:performance}, we find that our proposed approach achieves the best performance in terms of all evaluation metrics for both corpora.  In the previous studies, E2E SLU is difficult to beat the cascaded SLU, which can leverage large module-specific data to optimize individual modules. Our approach can unify well-optimized modules into a single model and further optimize the model with a unified objective of SLU.

\begin{figure}[htb]
\begin{minipage}[b]{1.0\linewidth}
  \centering
  \centerline{\includegraphics[width=6cm]{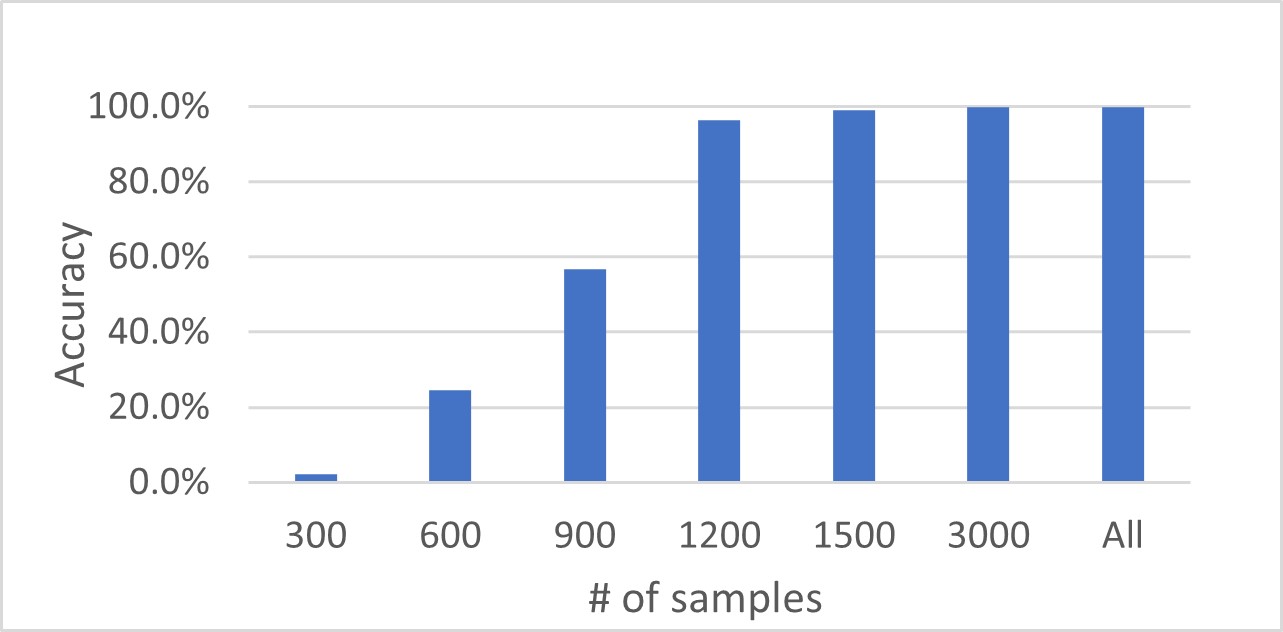}}
\end{minipage}
\caption{The number of training samples from FSC corpus vs. the intent accuracy by using the approach of E2E SLP one-step ($ASR\cup SLU$) }
\label{fig:numofsamples}
\end{figure}

The size of training data in the FSC corpus is relatively larger than other commonly used SLU corpora. We investigate whether we can reduce the number of utterances for transfer learning with the pre-trained SLP. The Figure~\ref{fig:numofsamples} illustrates the intent accuracy along with the number of training samples by using our proposed approach, i.e., E2E SLP one-step ($ASR\cup SLU$).  It shows that using only 3,000 samples by a random selection from the training set can reach a similar performance to that of using all 23,132 training samples.

\section{CONCLUSIONS}
\label{sec:conclusions}

The cascaded SLU can leverage ASR and NLU modules optimized with a large amount of module-specific training data and deliver a decent performance. It is a challenge for E2E SLU to beat it due to a lack of labeled data. We propose to leverage the pre-trained speech embeddings from ASR and the semantic representations from BERT and jointly optimize them for E2E SLU. We have demonstrated the effectiveness of our approach on FSC and ATIS datasets in this study. In the future, we will extend our method to multi-domain and multi-lingual SLU corpora and evaluate its generalization capability.




\bibliographystyle{IEEEbib}
\bibliography{strings,refs}

\end{document}